\documentclass[sigconf, 10pt, letter]{acmart}

\renewcommand\footnotetextcopyrightpermission[1]{} 
\usepackage{comment}
\usepackage{textcomp}
\usepackage{courier}
\usepackage{amsmath,amssymb,amsfonts}
\usepackage{algorithmic}
\usepackage{graphicx}
\usepackage{xcolor}
\usepackage{enumitem}
\usepackage{subfigure}
\usepackage{url}
\usepackage{hyperref}

\def\BibTeX{{\rm B\kern-.05em{\sc i\kern-.025em b}\kern-.08em
    T\kern-.1667em\lower.7ex\hbox{E}\kern-.125emX}}

\newcommand{\convince}{\texttt{CONVINCE}}

\usepackage{booktabs} 
\usepackage{paralist}

\usepackage[ruled]{algorithm2e}

\usepackage[colorinlistoftodos]{todonotes}

\begin{document}

\title{CONVINCE: Collaborative Cross-Camera \\ Video Analytics at the Edge} 
\author{Hannaneh Barahouei Pasandi, Tamer Nadeem}

\affiliation{%
  \institution{Dept. of Computer Science, Virginia Commonwealth University, Richmond, VA 23284, USA}
}
\email{barahoueipash, tnadeem@vcu.edu}

\begin{abstract}
Today, video cameras are deployed in dense for monitoring physical places e.g., city, industrial, or agricultural sites. In the current systems, each camera node sends its feed to a cloud server individually. However, this approach suffers from several hurdles including higher computation cost, large bandwidth requirement for analyzing the enormous data, and privacy concerns. In dense deployment, video nodes typically demonstrate a significant spatio-temporal correlation. To overcome these obstacles in current approaches, this paper introduces CONVINCE, a new approach to look at the network cameras as a collective entity that enables collaborative video analytics pipeline among cameras. CONVINCE aims at 1) reducing the computation cost and bandwidth requirements by leveraging spatio-temporal correlations among cameras in eliminating redundant frames intelligently, and ii) improving vision algorithms' accuracy by enabling collaborative knowledge sharing among relevant cameras. Our results demonstrate that CONVINCE achieves an object identification accuracy of $\sim$91\%, by transmitting only about $\sim$25\% of all the recorded frames.
\end{abstract}


\maketitle

\section{Introduction}
Driven by drastic fall in camera cost and the recent advances in computer vision-based video inference, organizations are deploying cameras in dense for different applications ranging from monitoring industrial or agricultural sites to retail planning. As an example, Amazon Go~\cite{amazongo} features an array of 100 cameras per store to track the items and the shoppers. Processing video feeds from such large deployments, however, requires a considerable investment in compute hardware or cloud resources. Due to the high demand for computation and storage resources, Deep Neural Networks (DNNs), the core mechanisms in video analytics, are often deployed in the cloud. Therefore, nowadays, video analytics is typically done using a cloud-centered approach where data is passed to a central processor with high computational power. However, this approach introduces several key issues. In particular, executing DNNs inference in the cloud, especially for real-time video analysis, often results in high bandwidth consumption, higher latency, reliability issues, and privacy concerns. Therefore, the high computation and storage requirements of DNNs disrupt their usefulness for local video processing applications in low-cost devices. Hence, it is infeasible to deploy current DNNs into many devices with low-cost, low-power processors. Worst yet, today video feeds are independently analyzed. Meaning, each camera sends its feed to the cloud individually regardless of considering to share possible valuable information with neighbor cameras and to utilize spatio-temporal redundancies between the feeds. As a result, the required computation to process the videos can grow significantly.

Motivated by the aforementioned hurdles, we believe that there is a need for a new paradigm that can benefit the current systems by lowering energy consumption, bandwidth overheads, and latency, as well as providing higher accuracy and ensuring better privacy by pushing the video analytics at the edge. We are \textit{convinced} that by looking at a network of cameras as a collective entity that leverages i) spatio-temporal collations among cameras in one hand, and ii) knowledge sharing (e.g., sharing input, intermediate state, or output of the DNN models) among relevant cameras in the other, we can utilize the aforementioned benefits in our systems. Prior works fail in addressing the challenge of large scale camera deployments where the compute cost grows exponentially by the increase in the number of deployed cameras. Most of the recent works only focus on a \textit{single} camera (not a collection of cameras) to perform the given vision task. Recent systems have improved analytics of live videos by using frame sampling and filtering to discard frames~\cite{ganesh:2018, chameleon:2018, awstream:2018}. However, the focus of these works are on optimizing the analytics overhead for individual video feeds.  

Our prior work~\cite{paradigm:2019} describes our vision of pushing video analytics to the network edge to leverage knowledge sharing and spatio-temporal correlation among nodes. To demonstrate the new opportunities and challenges in our vision, we have designed a centralized collaborative cross-camera video analytics system at the edge hereafter \convince\ \footnote{\footnotesize{CONVINCE: \underline{CO}llaborative \underline{\textcolor{gray}{N}} Cross-Camera \underline{VI}deo a\underline{N}alyt\underline{C}is at the \underline{E}dge.}} that leverages spatio-temporal correlations by eliminating redundant frames in order to reduce the bandwidth and processing cost, as well as leveraging knowledge sharing across cameras to improve the vision model accuracy. Applications that could benefit from such a system include, but not limited to, public and pedestrian safety, retail stores (e.g., Amazon Go) and vehicle tracking.

\noindent\textbf{Contributions.} This paper make the following contributions:

We propose \convince, a novel centralized video analytics framework that leverages cross-camera spatio-temporal correlations and knowledge sharing to reduce compute and bandwidth requirements while preserving privacy [\S\ref{sec:implementation}].
    
Evaluation using a Deep Learning (DL)-based object detection model (YOLO) shows that \convince\ can reduce the number of transmitted frames while preserving model accuracy~[\S\ref{sec:implementation}].

We identify the current challenges associated with \convince\ [\S\ref{subsec:challenges}]. We also explain how to extend \convince\ to be able to run on real-world deployment. For example, to achieve privacy, we explain how we can enhance \convince\ with a locally distributed learning mechanism using federated learning to update the shared vision model at the edge and preventing sensitive data to be sent to the cloud~[\S\ref{subsec:future-work}].

\begin{figure}[!t]
	\centering
	\includegraphics[width=0.48\textwidth,height=2.2 in]{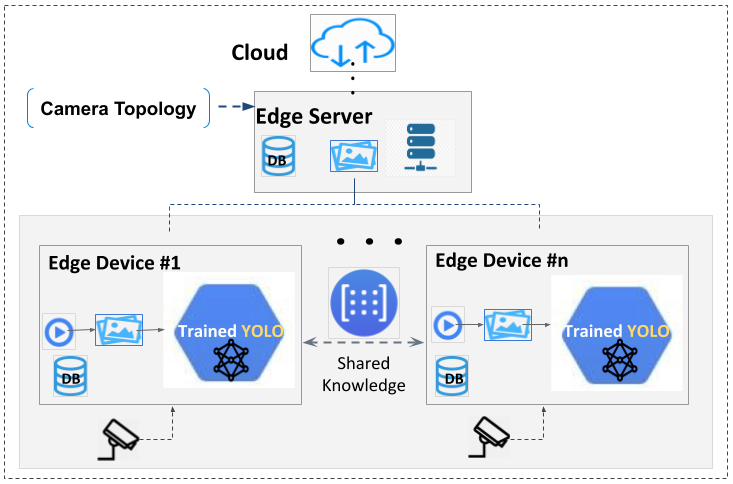}
	\caption{A Bird's-Eye View of \convince\ Framework } 
	\label{fig:eagle}

\end{figure}

\section{Framework Design Consideration}

To improve the current cloud-centered video analytics systems where each camera node transmits its feed in isolation, we have envisioned a novel paradigm to push the video analytics to the network edge to enable near real-time video analytics, lower latency, and to address privacy concerns. To do so, we proposed to look at a network of cameras at the edge as a collective entity that leverages spatio-temporal correlations among cameras and enables knowledge sharing across cameras.

The proposed paradigm can be realized either in a centralized or distributed setting. In both design approaches, we assume that each camera node has an embedded processing unit that is able to run moderate DNN models. The recent wave in "AI cameras" with embedded compute and storage resources~\cite{amazon} makes this assumption realistic. In the following, we describe both system design choices and their associated assumptions, requirements, and challenges.

\textbf{Centralized approach} In this approach, we assume that there is a centralized edge server that is powered with compute and storage requirements for a deeper video processing task. This server also keeps track of spatio-temporal correlations among cameras. In this setting, all the other camera nodes are connected (e.g., through Wi-Fi) to the edge server and transmit their pre-processed frames for further processing. The server is also aware of the camera network topology. We have implemented the centralized system architecture in this work (see Figure~\ref{fig:eagle}). Further details are provided in Section~\ref{sec:implementation}.

\textbf{Distributed approach} In this setting, all AI cameras are communicating with each other. Each camera needs to know the topology or the spatio-temporal correlations of the other cameras in the network. There are several key design challenges associated with a distributed setting that we highlight two of them in the following. 
One of the main design challenges in a this approach is the communication strategy among camera nodes in order to collectively improve their performance. The communication mechanisms between multiple nodes could be mainly categorized into two key approaches; individual peer-to-peer channels and all-to-all channels. Peer-to-peer channels will enable peer cameras to experience similar conditions (e.g., proximate cameras with overlapped Field of Views (FoV)) and targeting similar objectives to exchange their information with each other to speed-up the analytics process considering time constraints. Another challenge is to carefully design a trade-off between the information transmission time among the cameras and the delay in the inference step of running DNN models.

\section{Implementation and EVALUATION}
\label{sec:implementation}
Figure~\ref{fig:eagle} illustrates the overview of \convince\ framework. In this framework, each camera feed is processed locally to eliminate redundant frames. The output of the device-based processing is then sent to the edge server to evaluate the accuracy of the model running in each camera. In addition, \convince\ enables cameras to communicate with each other and share their obtained knowledge with potential peer cameras (see Subsection~\ref{results}).

\textbf{Experiment setup}
To evaluate the performance of

\begin{figure}[!t]
	\centering
	\includegraphics[width=0.35\textwidth,height=1.5 in]{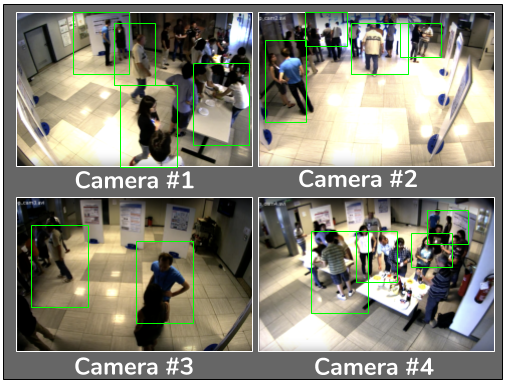}
	\caption{Different Camera Views in SALSA Dataset~\cite{salsa}} 
	\label{fig:salsaview}
\end{figure}

~\convince, we have used YOLO-v2 (You Only Look Once)~\cite{yolo} which is a CNN  for object detection. The object detection task consists in determining the \textit{location} on the frame where certain objects are present, as well as \textit{classifying} those objects. The output of YOLO is a vector of bounding boxes and class predictions for the objects in the bounding boxes. YOLO-v2 has a total of 32 convolution layers that make it suitable for low resource devices. For evaluation, we used YOLO-v2 values trained on the VOC2012 dataset~\cite{voc}.  \footnote{Due to space limitation, we refer the interested reader about YOLO to~\cite{yolo}.} We evaluated \convince\ using YOLO-v2 for people counting task on SALSA dataset~\cite{salsa} which contains footage from four cameras placed on four corners of an indoor area (see Figure~\ref{fig:salsaview}), with significant FoV overlap among the cameras. It contains uninterrupted recordings of an indoor social event involving 18 subjects over 60 minutes. The reason for our choice of this dataset is to analyze the potential of using spatio-temporal correlations of cross-camera video analytics. Depending on what environment (indoor, outdoor) and the objective of deployed cameras (e.g., security cameras, people counting outdoor), the number of detected objects in a frame for a certain time window does not change often based on the observation in~\cite{jain:2019scaling}. Motivated by such observations, \convince\ edge devices only transmit those frames in which a new object is detected to the edge server. Our own observation on SALSA dataset also suggests that such observations are likely to happen in typical in camera deployments. In \convince\, a newly detected object means there is a \textit{new bounding box} detected in a sequence of frames.

\textbf{Performance metrics} Measuring performance is a trade-off between the model accuracy, resource efficiency, and performance cost optimization of data analysis on resource-challenged camera nodes. One of the objectives of \convince\ is to reduce the number of redundant frames to save the network bandwidth and processing time of the video analytics while maintaining the model accuracy. We measured \convince\ performance by i) people counting accuracy,
and b) fraction of transmitted frames by cameras.

\subsection{Results}
\label{results}

\begin{figure}
\centering
\subfigure[]{\label{single}\includegraphics[width=41mm]{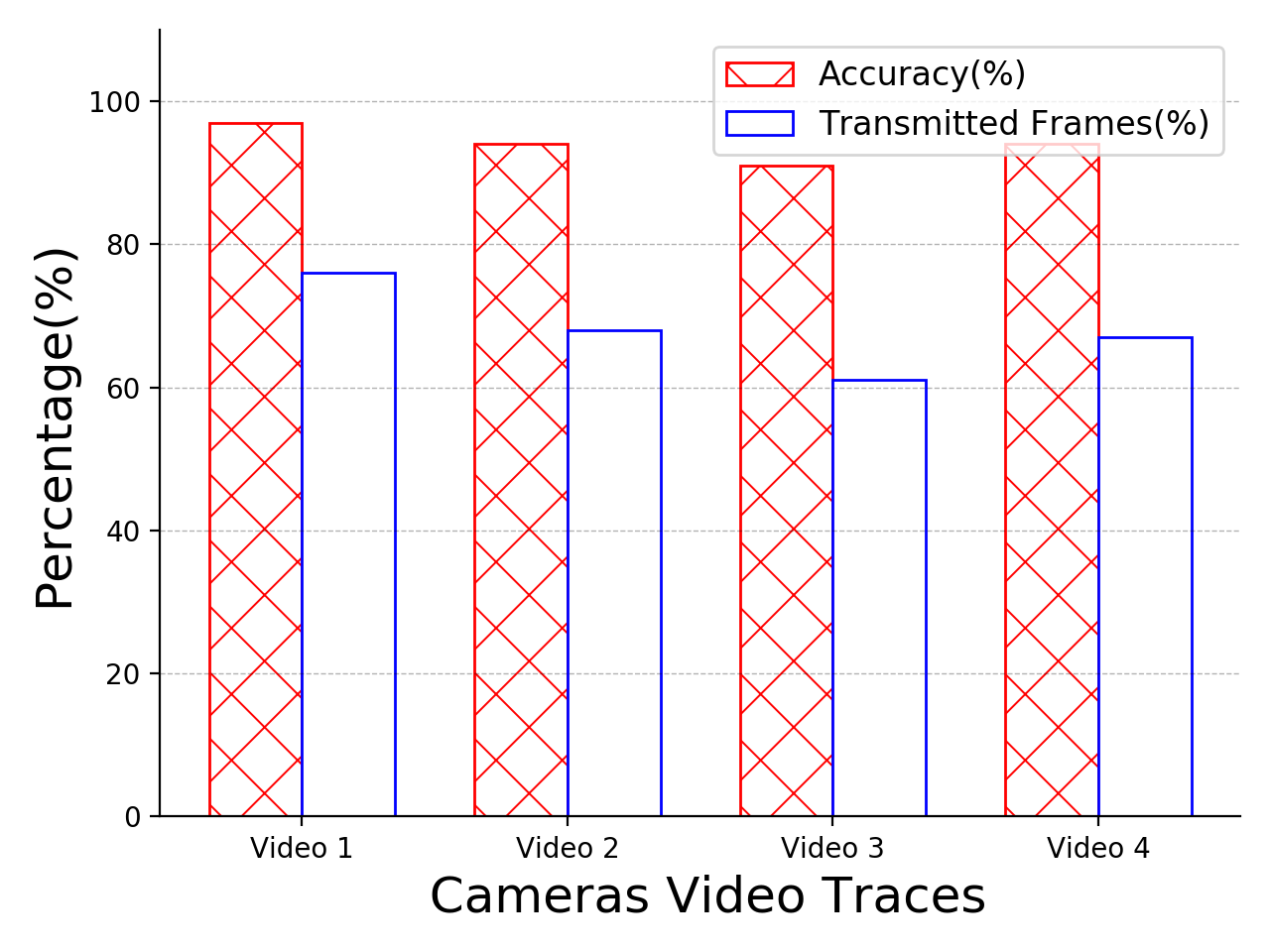}}
\subfigure[]{\label{collaborative}\includegraphics[width=41mm]{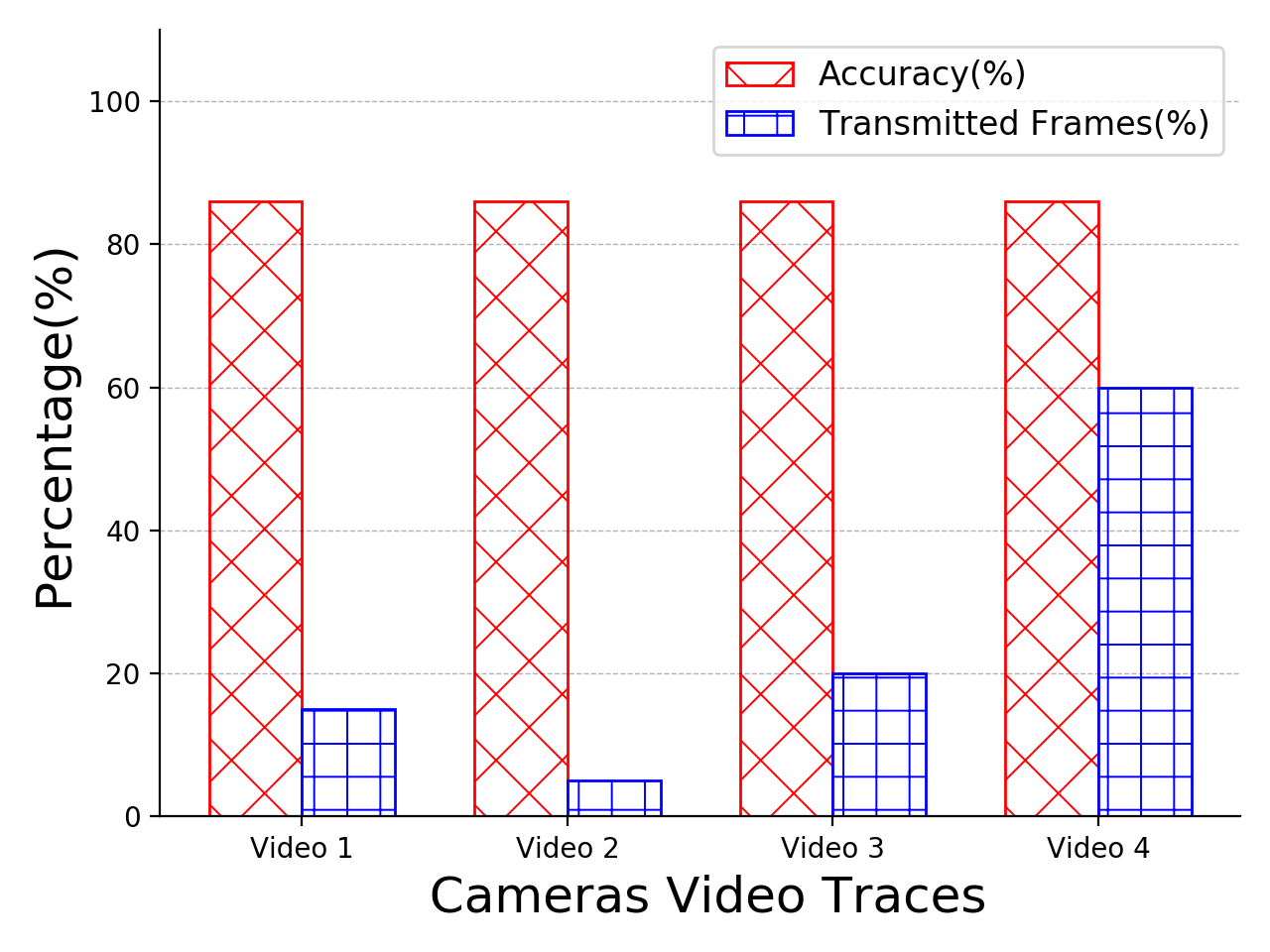}}
\caption{ Performance evaluation of~\subref{single} Single camera frame transmission~\subref{collaborative} Collaborative cross-camera frame transmission.}
\label{th-compare}
\end{figure}

\textbf{Isolated camera frame transmission}
We evaluate the performance of \convince\ when individual cameras transmit only the frames with a newly detected object to the edge server. Edge server calculates the people counting accuracy using the transmitted frames by all cameras.
In this experiment, we have evaluated \convince\ for people counting task on four recorded videos using SALSA dataset.  Figure~\ref{single} shows the performance of \convince\ for different recorded videos. As shown in the Figure, Although cameras transmit only selected frames which account on average for about $\sim$ 65\%  of their recorded frames to the edge server, people counting accuracy is maintained a value of about $\sim$ 94\% in all the four video traces. This experiment shows that by intelligently selecting only informative frames to be processed, we can eliminate redundant frames, which results in less bandwidth and processing consumption while preserving the model's accuracy.

\textbf {Collaborative cross-camera frame transmission}
In this experiment, we take a further step to evaluate the performance of \convince\ in a cross-camera setting where there is a significant overlap in cameras FoV. The purpose of this experiment is to examine the fraction of transmitted frames with significant FoV overlap which are analyzed only once and to see whether \convince\ accuracy will be compromised or not. For this purpose, we select one of the cameras as a \textit{supreme} camera ( namely Camera \#4). The supreme camera is chosen based on the higher accuracy in people counting task it achieves compared to the other cameras and their transmitted frames. As shown in Figure~\ref{collaborative} cameras on average transmit only  $\sim$ 25\% of their total frames. Although viewing cameras collectively decreases the total number of transmitted frames from  $\sim$ 60\% in the previous experiment to only $\sim$ 25\% the accuracy of the model also drops to $\sim$ 86\%. Therefore, we need sophisticated mechanisms in which the trade-off between model accuracy, resource efficiency, and cost of data analysis is carefully considered.

\textbf{Collaborative cross-camera knowledge sharing}

 Cameras that are installed in vulnerable positions could suffer from low-quality images at different times of the day or under various weather conditions (e.g., under extreme luminance during the day or a rainy weather condition), which results in lower accuracy due to poor quality of input data. However, other cameras installed in a better position which have an overlapping FoVs would have better inference performance. Therefore, such cameras can complement each other via collaboratively sharing their inputs. In such a scenario, knowledge sharing among cameras may be as simple as sharing input frames among relevant cameras.  Knowledge sharing can also mean to share an intermediate state of the DNN model with other cameras running the same model to enhance their accuracy.
 
In \convince\  a non-collaborative mechanism means each camera runs the YOLO-v2 algorithm for people detection task individually. Inspired by~\cite{misra:2019}, we share the intermediate output of the YOLO algorithm which is the detected bounding boxes and their associated confidence level with relevant cameras. In YOLO-V2 there is a final step called Non-maximum suppression (NMS) which ensures finding a single bounding box for each detected object among several boxes detected for the same object. NMS ensures finding an optimal bounding box for each object while suppressing any detected box when the degree of overlap between the detected and the selected box is lower than the default threshold. In collaborative camera-setting, relevant collaborating cameras share their inference states before and after the NMS step with other cameras. The collaborator bounding boxes are transformed to the same coordinate system as the supreme camera and pairs of bounding boxes are matched using the Hungarian algorithm. Those bounding boxes that fall close within the same areas across cameras are assigned a higher confidence score weight since they are detected by multiple cameras.

In this experiment, the bounding box coordinates and their corresponding confidence scores with the other cameras are shared. The supreme camera (Camera \#4) is the baseline. As shown in Table~\ref{table-collaborative} frames transmitted  only by Camera $\#4$ result in  $\sim$67 \% accuracy. We then share the bounding box information of Camera $\#4$ with Camera $\#3$. This shared knowledge improves the accuracy of people counting to $\sim$89\%. As shown in Figure~\ref{collaborative}, frames transmitted by Camera $\#3$ and $\#4$ accumulate 90\% of all the transmitted frames. As expected, adding Camera $\#1$ and $\#2$ marginally enhances the accuracy to about $\sim$91\%. By sharing bounding boxes between cameras, we show that the accuracy of the model increased by about $\sim$5\% compared to the previous experiment where the supreme camera does not share the intermediate model information with other cameras.

Our early results demonstrate the opportunities that 

\convince\ can bring to the current video analytics systems. However, there are many challenges that need to be addressed. The results could be highly depend on the position, angle, overlapped FoVs' of cameras. Therefore, for different camera settings we need coping mechanism.  For example, a dense camera deployment may require a clustring algorithm to group relevant cameras together based on their position and FoVs. In the following section, we discuss some of the challenges and future directions of our approach.

\begin{table}

\centering
\caption{Comparing people counting accuracy using different camera frames.  \label{table-collaborative}}
\begin{tabular}{|c|c|}
\hline
\textbf{Camera feed used}      & \textbf{Accuracy(\%)} \\ \hline
Camera \#4 (Baseline)                     & $\sim$67        \\ \hline
Camera  \#3, 4                & $\sim$89        \\ \hline
Camera \#1, 2, 3, 4              & $\sim$91                     \\ \hline
\end{tabular}

\end{table}

\section{Discussion}
\subsection{Current Challenges}
\label{subsec:challenges}
\textbf{Dealing with adversarial nodes}
As mentioned in~\cite{collaborative:2019}, the performance and accuracy of our system could be affected considered with the presence of adversarial cameras. This is because all nodes including adversarial cameras share their inference with the proximity nodes. Therefore, we need sophisticated mechanisms such as centralized trust management~\cite{centralizedTrust:2017} in the edge server to be able to calculate appropriate trust scores for each camera based on the feedback provided to the edge server. The trust score could have a range between [0,1]. Table~\ref{tab:indexing} provides an example of such a calculated scoring mechanism. Such trust mechanisms can become extremely important in military scenarios where nodes may not necessarily trust each other.

\begin{table}[]
\footnotesize
\caption{Trust Value Indexing example by the centralized edge server \label{tab:indexing} }
\label{tab:my-table}
\begin{tabular}{|c|c|c|}
\hline
\textbf{Trust Score} & \textbf{Description}     & \textbf{Label}    \\ \hline
0                    & Completely untrustworthy & Extremely harmful \\ \hline
0.3                  & Risk trust               & Risky             \\ \hline
0.5                  & Semi-trust               & Semi-Safe         \\ \hline
0.7                  & Trustworthy              & Safe              \\ \hline
1.0                  & Completely Trustworthy   & Completely Safe   \\ \hline
\end{tabular}
\end{table}

\textbf{Dealing with small training datasets}
The term \textit{learning more from less} also applies in the machine learning domain when there are only a few samples to learn from. In camera deployment setting, sometimes a camera is installed where it does not receive many useful samples to learn from. For instance, a camera that is installed in the main hallway may detect many samples, while another camera detects very few at the same time. One approach to overcome this challenge could be to share the samples of the camera with more samples with the other camera to train its model without sacrificing privacy. We could also use techniques such as few-shot learning where we have only a few examples of sample data to learn from. Few-shot learning methods can be roughly categorized into two classes: data augmentation and task-based meta-learning. For example, in~\cite{few-shot-learning:2019} the proposed model gave state-of-the-art results and paved the path for more sophisticated meta-transfer learning methods.

\subsection{Future Work}
\label{subsec:future-work}
\textbf{Designing a module to model spatio-temporal correlations}
Cross-camera movements (e.g., people or traffic) demonstrates a high degree of spatial and temporal correlation. A movement between two cameras could be defined as the set of unique objects detected in the first camera that are then detected in the second camera.
Exploiting spatio-temporal correlations, by itself potentially saves compute resources. In \convince\ we need mechanisms to capture and model such correlations of detected objects between pairs of cameras' views.

\textbf{Identifying collaborative nodes} In our experiment, we have only used four cameras that are calibrated. Thus, it was easy to identify the collaborative nodes using trial and error. In a real-world setting, the number of cameras can be as large as an array of hundred cameras e.g., in Amazon Go stores. Sometimes cameras are not stationary which can lead to the change of ideal collaborators. Therefore, we need sophisticated mechanisms to identify potential peer collaborators. One possible approach would be to cluster cameras with spatial or temporal correlations and select a supreme camera for the cluster based on the area of coverage and resolution quality provided by the camera.

\textbf{Collaborative privacy-preserving video analytics}
The use of computer vision technologies is not limited to the rapid adoption of facial recognition technologies but is also extended to facial expression recognition, scene recognition, etc. These developments raise privacy concerns regarding the collection and the use of sensitive data. These concerns can grow to the extent that regulators and authorities take serious actions about these technologies. Most of the current privacy-aware video streaming approaches involve denaturing, which means the content of images or video frames is modified based on a guided privacy policy. In addition, cheap internet-of-visual-things (IoVT) along with emergent of vision processing technologies made video recording and sharing more attractive. The cycle-consistent GAN~\cite{CycleGAN:2017} is a popular technique to transform a video frame from one style to another and received significant attention in the literature. Some of the recent efforts~\cite{UPRID:2020} propose CycleGAN for person re-identification task by using unsupervised classification methods~\cite{wickramasinghe:2019}. However, in practice, there are two privacy/security issues to be addressed before deploying cycleGANs on to the edge devices. Firstly, any adversary can recover the original contents of cycGAN-transformed video frames if it can access the same cycGAN-enabled edge device for training data collection. Secondly, it is not easy to verify that the transformed frame is legitimate or not as shown by a technique in~\cite{steganography:2017}. As a coping mechanism to such issues, authors in~\cite{secgan:2019} propose to append a watermark to each input in the training phase that is treated as a secure key to reduce the cycleGAN shortcomings in terms of privacy. However, all the stenography-based approaches typically consider a non-collaborative setting.
In \convince\ centralized approach, privacy can be achieved by using techniques such as a modified version of federated learning. It allows for a distributed training scheme where first each device is initialized by a single model e.g., object detection. When a new object is detected it updated its local model. It then sends an update (model parameters and corresponding weights of non-sensitive data) to the edge server. The update is then averaged over all other edge nodes' updates to improve the shared model. Therefore, there is no privacy breaches between nodes even if one of the cameras is compromised.  

\section{CONCLUSION}
This paper describes \convince\ a collaborative intelligent cross-camera video analytics at the edge, a paradigm where video nodes perform vision tasks collaboratively on resource-constrained cameras on the network edge. We believe that such intelligent cross-camera collaboration can significantly lower energy, bandwidth overheads and latency, and provide better accuracy while ensuring privacy. Our early results highlight the benefits of such a paradigm that can bring to the current systems. We have also discussed some of the key challenges and future directions in realizing the proposed paradigm. Although we only focused on collaborative cross-camera video analytics application, we believe the proposed collaborative paradigm could be applied to other types of IoT devices/sensors.

\bibliographystyle{ACM-Reference-Format}
\bibliography{references}

\end{document}